\documentclass[conference]{IEEEtran}
\IEEEoverridecommandlockouts

\usepackage[cmex10]{amsmath}%American Math Society(AMS) math formatting
\usepackage{amssymb,amsfonts}%AMS extra symbols and fonts
\usepackage{dblfloatfix}%fix double column figure ordering and placement

\usepackage[ruled,vlined]{algorithm2e}
\usepackage{graphicx}
\graphicspath{{Figures/PDF/}{Figures/PNG/}}

\usepackage{booktabs}
\usepackage{siunitx}
\usepackage[numbers,compress]{natbib}
\usepackage{texnames}
\usepackage{bm,bbm}
\usepackage{soul}
\usepackage[table]{xcolor}
\usepackage{enumitem}

\usepackage[most]{tcolorbox}

\newtcolorbox{highlightparagraph}{
    enhanced,
    breakable,
    colback=yellow!15,
    colframe=yellow!50!black,
    boxrule=0.4pt,
    arc=1mm,
    left=2mm,
    right=2mm,
    top=1mm,
    bottom=1mm,
    before skip=2mm,
    after skip=2mm
}

\begin{document}

% \title{Entropy-Centric Explainable AI for Remote Sensing Image Segmentation}

% \author{\IEEEauthorblockA{Ali Saleh\IEEEauthorrefmark{1}\IEEEauthorrefmark{3},
% Abdul Karim Gizzini\IEEEauthorrefmark{2},
% Mohamad Ghassany\IEEEauthorrefmark{3},
% Ali J. Ghandour\IEEEauthorrefmark{4} % ,
% }

% \IEEEauthorblockA{
% \IEEEauthorrefmark{1} Faculty of Engineering, Lebanese University, Beirut Lebanon.\\
% \IEEEauthorrefmark{1} University of Paris-Est Créteil (UPEC), LISSI/TincNET, F-94400, Vitry-sur-Seine, France. \\
% \IEEEauthorrefmark{3} EFREI, EFREI Research Lab, F-94800 Villejuif, France.\\
% \IEEEauthorrefmark{4} National Center for Remote Sensing - CNRS, Mansourieh, 22411, Lebanon.\\
% Corresponding Author: aghandour@cnrs.edu.lb}}

\title{Entropy-Centric Explainable AI for Remote Sensing Image Segmentation\\
\author{\IEEEauthorblockN{Ali Saleh\IEEEauthorrefmark{1}\IEEEauthorrefmark{3}, 
Abdul Karim Gizzini\IEEEauthorrefmark{2}, 
Mohamad Ghassany\IEEEauthorrefmark{1}, 
Ali J. Ghandour\IEEEauthorrefmark{3}
} 
\IEEEauthorblockA{\IEEEauthorrefmark{1} Faculty of Engineering, Lebanese University, Beirut, Lebanon}
\IEEEauthorblockA{\IEEEauthorrefmark{2}University of Paris-Est Créteil (UPEC), LISSI/TincNET, F-94400, Vitry-sur-Seine, France.} 
\IEEEauthorblockA{\IEEEauthorrefmark{3} EFREI, EFREI Research Lab, F-94800 Villejuif, France.\\}
\IEEEauthorblockA{\IEEEauthorrefmark{4}National Center for Remote Sensing - CNRS, Mansourieh, 22411, Lebanon.\\
Corresponding author: aghandour@cnrs.edu.lb}
    }
}

\maketitle

\begin{abstract}
Artificial intelligence (AI) has become a powerful approach to solving complex problems in critical domains. Many concerns arise regarding the decision-making process of its models, which is mainly due to deep neural networks outperforming their peers at the cost of ambiguity in feature extraction and prediction. Consequently, in critical domains like remote sensing, where we need to analyze high-resolution imagery using black-box models, the lack of transparency limits their trust and thus their adoption. In front of this reality, explaining and understanding the complex AI model's decision-making process becomes a must. Explainable AI (XAI) aims to bridge this gap by providing insights into how and why certain decisions are made. While significant progress has been achieved in explaining image classification tasks, image segmentation still offers considerable room for improvement. In this context, this paper proposes an entropy-centric XAI method for semantic segmentation. Moreover, a new XAI evaluation methodology is proposed to efficiently measure the relevance of the highlighted regions by the proposed XAI method. Experimental results demonstrate the superiority of the proposed XAI method in comparison to the recently adapted XAI methods for semantic segmentation.

\end{abstract}

\begin{IEEEkeywords}
	Explainable AI, Remote Sensing, Image Segmenation, Entropy-Centric.
\end{IEEEkeywords}

\section{Introduction}

Artificial Intelligence (AI) has achieved promising success in various domains, including remote sensing, where it is employed in critical applications, such as land use monitoring \cite{SANGEETHA2025455}, environmental assessment \cite{RCSANTOS2025123864, electronics14040696}, and disaster management \cite{article_1580460}. AI-based solutions introduced automated feature extraction compared to traditional machine learning schemes. Feature extraction is one of the most complex and time-consuming phases in the training process for any model. The lack of transparency in extracting the desired features results in losing the interpretability of the decision-making methodology employed by the corresponding model. This raises concerns that impact human trust in AI, including its fairness and the ethical aspects of its decisions. In this context, the emergence of Explainable Artificial Intelligence (XAI), a field dedicated to uncovering the reasoning behind model predictions, has become a must. Hence, providing explanations that allow humans to understand, validate, and trust AI-driven decisions.

XAI schemes can be applied to different applications, including image classification and semantic segmentation. Both of these applications can benefit from XAI in revealing the key image features that mostly influenced the model prediction, enabling more interpretability and thus responsible use of AI in remote sensing. Despite the XAI success achieved in image classification, extending this success to image segmentation is challenging due to the spatial correlation between the pixels and regions. Among different XAI methods~\cite{speith2022review}, attribution methods focus on assigning importance scores to the input features, indicating their contribution to a model prediction. Thus, understanding what features are affecting its decisions~\cite{abhishek2022attributionbasedxaimethodscomputer}. XAI attribution methods are further divided into different classes: (\textit{i}) Gradient-based, where the importance score is computed based
on the gradients of backpropagation, (\textit{ii}) Sampling-based methods that treat the model as a black box and systematically sample
parts of the input and observe output changes to estimate
feature importance. It is worth mentioning that evaluating the robustness and efficiency of different XAI methods is essential for validating their faithfulness and reliability~\cite{shreim2023trainable}.

\begin{figure*}[t]
  \makebox[\linewidth][c]{%
    \includegraphics[width=1.15
    \linewidth]{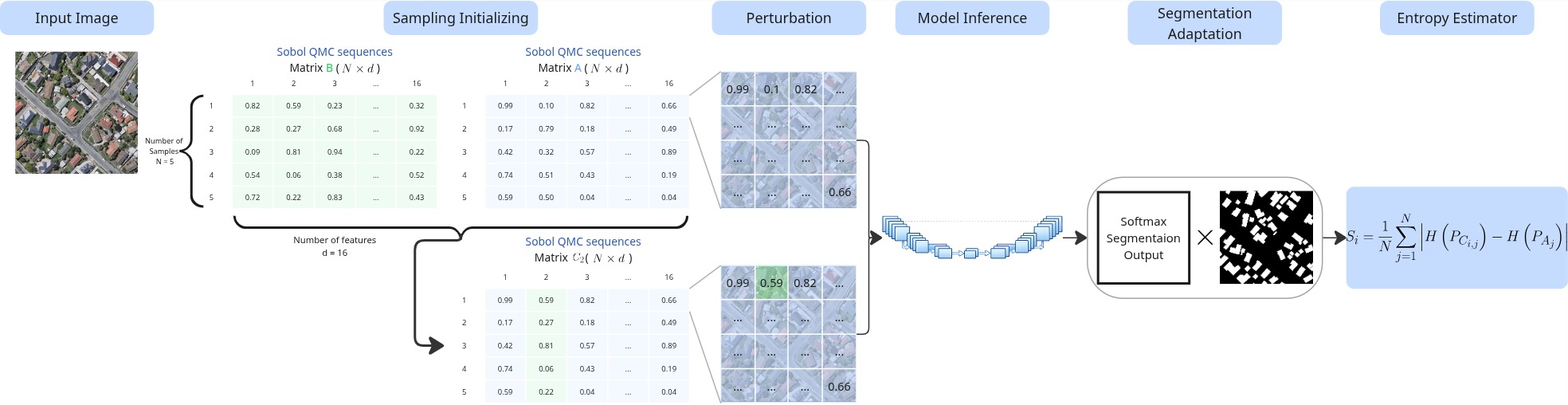}
  }
  \caption{Propsoed Entropy-centric XAI Method.}
  \label{fig:MethodsWorkPipeline}
\end{figure*}

In this context, this work focuses on proposing a sampling-based XAI method based on the entropy uncertainty principle. It consists of 2 sampling phases: Creation of Base sampling matrices then the perturbation one for input image perturbation, followed by a model inference to compute the target object entropy spatial scores to determine the importance scores for each reagion in the image.  Moreover, to efficiently evaluate the performance of the proposed Entropy-Centric XAI method, we propose a new XAI evaluation methodology that primarily
assesses whether highlighted relevant regions outside the target object are truly influential. The performance evaluation performed using the WHU dataset for building footprint segmentation~\cite{model} shows the superiority of our proposed entropy-centric method compared to the recent XAI methods adapted for semantic segmentation~\cite{gizzini2023extending}.

To sum up, the contributions of this paper can be listed as follows:

\begin{itemize}
    \item Proposing Entropy-Centric, a sampling XAI method for semantic segmentation using the entropy uncertainty methodology. 

    \item Proposing an XAI performance evaluation methodology that utilizes three metrics: (\textit{i}) prediction confidence score, (\textit{ii}) IoU score, and (\textit{iii}) Entropy score. 
    
    \item Performance evaluation using the WHU dataset for building footprint segmentation~\cite{8444434}.
\end{itemize}

The remainder of this paper is organized as follows: Section II presents the proposed entropy-centric XAI methodology. The performance of the benchmarked XAI methods in terms of the proposed XAI evaluation methodology is analyzed and discussed in Sections III and IV, respectively.

\section{Conventional Sobol XAI Method}

% This section highlights the main principles of the proposed entropy-centric XAI method, as well as the proposed XAI evaluation methodology. First of all, we breifly presents the XAI Sobol method, which motivates our proposed method.

% \subsection{Conventional Sobol Method}

Sobol indices~\cite{fel2021sobol} is a variance-based global sensitivity method that measures the contribution of input variables to the output variance of a model. For a model output $\mathbf{f(x)}$ with input features $\mathbf{x=\{x_1,\dots,x_n\}}$, the total variance $D$ is decomposed into main and interaction effects:

\begin{align}
D &= \sum_{i=1}^n D_i + \sum_{i<j} D_{ij} + \cdots + D_{1,2,\dots,n},
\end{align}

where $D_i$ denotes the variance contribution of input $i$ and higher-order terms represent interactions. The total Sobol index quantifying the importance of input $i$, including all its interactions, is defined as:

\begin{align}
S_{T_i} = 1 - \frac{\mathrm{Var}_{x_{\sim i}}\!\left[\mathrm{E}_{x_i}\!\left[f(x \mid x_{\sim i})\right]\right]}{D},
\end{align}

where $x_{\sim i}$ denotes all the inputs except $i$. This index quantifies the importance of each feature in its interaction with other inputs on the model output variability.\\

Sobol indices work by perturbing inputs systematically using predefined sampling masks according to the quasi-Monte Carlo sampling technique \cite{fel2021sobol}. Unlike purely binary random sampling masks, QMC uses low-discrepancy sequences that ensure better coverage of the input space and accelerate the convergence of sensitivity estimations.

In the conventional Sobol method, calculating total-order indices $S_{T}$ requires extensive model evaluations over a large number of perturbation samples, which quickly becomes computationally prohibitive. Here Jansen estimator \cite{fel2021sobol} is used primarily to address the high computational cost associated with estimating variance-based sensitivity indices, especially in high-dimensional inputs such as images. Its adoption is motivated by the need for an efficient, stable, and low-variance estimator that can provide accurate sensitivity indices with fewer model evaluations. Specifically, the Jansen estimator calculates total Sobol indices $\hat{S}_{T_i}$ using paired sets of perturbation masks—matrices $A$ and $B$, the base sampling masks matrices initialized using QMC, and $C_i$, where $C_i$ replaces the $i^{th}$ column of $A$ with the corresponding column from $B$. This paired sampling allows the estimator to approximate the variance contributions of each input dimension more efficiently. The estimation formula is given by:

% Sobol indices are estimated by systematically perturbing inputs using quasi-Monte Carlo (QMC) sampling~\cite{fel2021sobol}, which improves coverage of the input space and convergence. To reduce the high computational cost of estimating total-order indices in high-dimensional settings, the Jansen estimator~\cite{fel2021sobol} is employed. It computes $\hat{S}_{T_i}$ using paired sampling matrices $A$ and $C_i$ as:

\begin{align}
\hat{S}_{T_i} = \frac{\frac{1}{2N} \sum_{j=1}^{N} \left(f(\mathbf{A}_j) - f(\mathbf{C}_{i,j})\right)^2}{\hat{V}},
\end{align}

where $\hat{V}$ is the empirical variance of the model outputs based on samples from $A$. The combination of the Jansen estimator and QMC sampling allows the calculation of Sobol Indices in problems with high-dimensional outputs, such as images, with a smaller number of forward passes while maintaining estimation accuracy. We note that in our work, conventional Sobol is adapted to image segmentation to evaluate the influence of input image patches on the segmentation output variability, allowing for the identification of the critical regions.

\begin{figure*}[t]
    \centering
    \includegraphics[width=1\linewidth]{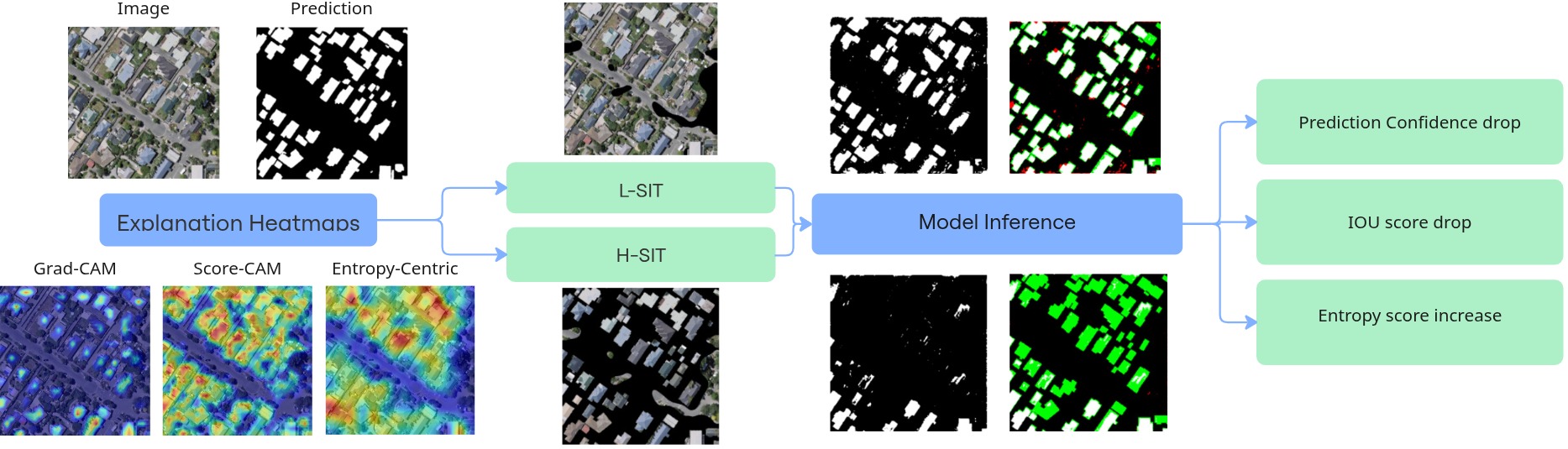}
    \caption{The evaluation framework proposed through Irrelative Validation and False Sailency Detection methods applied on 3 heatmaps generated by the XAI methods. The masked images, prediction and IOU maps are based on the Entropy-Centric heatmap.}
    \vspace{1em}
    \label{fig:evaluation-flowchart}
\end{figure*}

\section{Proposed Entropy-Centric XAI Method}

The proposed Entropy-Centric XAI method utilize Sobol XAI concept and is based on the entropy methodology that measures how the perturbation of different input patches affects the output entropy and hence identifies the regions that maintain confidence in the segmentation output. The greater the entropy change upon masking a region, the higher its attribution score. The binary entropy $H \in [0,1]^{H \times W}$ of the predicted target class probability $p_k$ at each pixel can be expressed as:
\begin{align}
H = -p_k \log(p_k) -(1-p_k)\log(1-p_k)
\label{eq:BinaryEntropy}
\end{align}

Through Entropy-Centric we adapt Sobol to image segmentation and continue with developing the entropy methodology. Adapting to image segmentation necessitates a fundamental shift from explaining scalar output values, as in classification, to handling spatially structured outputs. In segmentation tasks, the primary objective is to interpret the model’s predictions at the pixel or region level for specific target objects, rather than for the image as a whole. 

To achieve this, each perturbed image forward pass of the segmentation model yields a spatial score map $S_c \in \mathbb{R}^{H \times W}$, essentially a probability distribution indicating the likelihood of the object class at every pixel. The output map corresponding to the target object is isolated by multiplying the model’s output by the object’s mask $M \in [0,1]^{H\times W}$.
\begin{align}
    f_{masked}(S_c)=S_c \odot M
    \label{eq:MaskedOutput}
\end{align}

This allows for analyzing the contribution of input regions to the prediction of the relevant object, discarding spurious influences elsewhere in the image.

Figure \ref{fig:MethodsWorkPipeline} illustrates the Entropy-Centric pipeline. It begins by initializing the base sampling matrices $A$ and $B$ using quasi-Monte Carlo (QMC) methods~\cite{fel2021sobol}, followed by generating the perturbation masks $C_i$ based on these matrices. The perturbed images $A_j^{'}$ and $C_{i,j}^{'}$ are subsequently created by applying the upsampled $A_j$ and $C_{i,j}$ masks according to eq.~\ref{eq:Upsampling}.
\begin{align}
    A_j^{'}= I \odot \mathrm{up}(A_j)
    \hspace{2em}
    C_{i,j}^{'}= I \odot \mathrm{up}(C_{i,j}),
    \label{eq:Upsampling}
\end{align}
where $I\in \mathbb{R}^{H \times W}$ denotes the input image with dimesniosn $H$ and $W$. Next, model inference is performed to obtain the spatial score maps of the perturbed inputs $S_{A_j^{'}}$ and $S_{C_{i,j}^{'}}$. Then target masking them using eq.~\ref{eq:MaskedOutput} to get $f_{masked}(S_{A_j^{'}})$ and $f_{masked}(S_{C_{i,j}^{'}})$. The Softmax probabilities are then computed as: 
\begin{equation}
    \left\{
            \begin{array}{ll}
                 P_{A_{j}}=Softmax(f_{masked}(S_{A_j^{'}}))\\
                 P_{C_{i,j}}=Softmax(f_{masked}(S_{C_{i,j}^{'}}))
            \end{array}\right.
        \label{eq:EntropySoftmax}
\end{equation}

Finally, the importance scores are calculated, after calculating the entropy of the target class from its softmax probabilities through eq.\ref{eq:BinaryEntropy}, using :
\begin{align}
    S_i=\frac{1}{N}\sum_{j=1}^N \left| H \left( P_{C_{i,j}} \right) -  H\left( P_{A_{j}} \right) \right|
    \label{eq:Entropy-CentricEstimator}
\end{align}
where $H( C_j )$ and $H( A_j )$ denote the entropy values computed for perturbed and base samples, respectively. We note that the proposed Entropy-Centric XAI method focuses on the uncertainty %rather than only the output score, which offers a different explainability perspective.
which offers a different explainability approach than the Sobol sensitivity method by revealing areas degrading the certanity of the model after perturbation.

\begin{figure*}[t]
    \centering
    \includegraphics[width=0.55\linewidth]{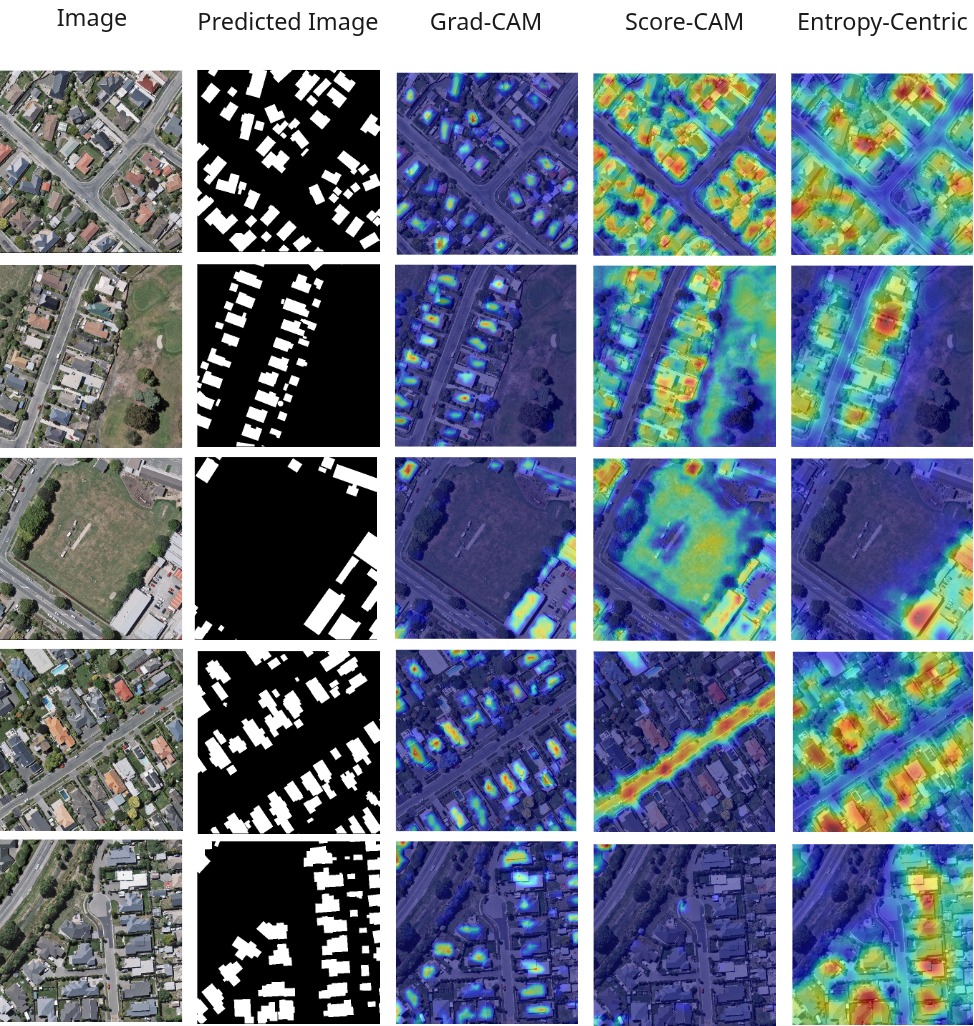}
    \caption{Sample of XAI heatmaps generated by the 3 methods Grad-CAM, Score-CAM and Entropy-Centric that summarizes all the cases ion the testing results}
    \vspace{1em}    
    \label{fig:heatmaps-results}
\end{figure*}

\subsection{Proposed Evaluation Methodology}

XAI performance evaluation remains challenging due to the absence of a unified evaluation framework, making fair comparison between XAI methods difficult. The authors in~\cite{gizzini2023extending} proposed an XAI evaluation methodology we refer to as Low-Salience Irrelevance Test (L-SIT) that evaluates whether regions assigned low importance in the explanation are indeed irrelevant to the model. L-SIT removes pixels with explanation values below a chosen threshold and measures the resulting change in prediction confidence and segmentation accuracy. Given the explanation heatmap $E$ with $\phi$ the set of highlighted pixels above a specific threshold, and $T$ the set of target object pixels of the input image $I$. The perturbed image by L-SIT can be expressed as:
\begin{equation}
    I^{\prime}_{\text{L-SIT}} = T \cup {\phi}.
\end{equation}

A small performance drop confirms the irrelevance of low-salience regions outside the target object, while a large drop reveals their hidden importance. However, L-SIT cannot detect cases where unimportant regions are mistakenly given high salience, which motivates the need for a complementary evaluation strategy.

To address this limitation, we propose a new XAI evaluation methodology, denoted as High-Salience Influence Test (H-SIT), which primarily assesses whether regions outside the target object but highlighted as important are truly influential. By removing these out-of-object, high-saliency regions, the resulting performance deterioration directly reflects their actual relevance to the model’s decision process. The perturbed image by H-SIT is defined as:

\begin{equation}
    I^{\prime}_{\text{H-SIT}} = T \cup \overline{\phi}.
\end{equation}
where $\overline{\phi}$ is the set of highlighted pixels below a specific threshold. Through the proposed H-SIT XAI evaluation methodology, a large drop in the considered metric is expected since the regions most affecting the model decision are removed, enabling the identification of falsely highlighted regions that L-SIT methodology alone cannot capture.

\begin{table*}[t]
    \renewcommand{\arraystretch}{1.6}
    \centering
    \caption[Quantitative Results]{L-SIT and H-SIT quantitative results for the three metrics: (\romannumeral 1) Mean of the Prediction Confidince Score Drop, (\romannumeral 2) Mean of the IOU Score Drop, (\romannumeral 3) and Mean of the Entropy Score Increase at 0.1 threshold over the WHU dataset.}

    \begin{tabular}{|c|cc|cc|cc|}
        \cmidrule(r){2-7}
        \multicolumn{1}{c}{} &
        \multicolumn{2}{|c|}{Prediction Confidence} &
        \multicolumn{2}{|c|}{IOU Score} &
        \multicolumn{2}{|c|}{Entropy Score} \\
        \cmidrule(r){1-7}
        \multicolumn{1}{|c}{XAI Methods} & 
        \multicolumn{1}{|c}{\shortstack{L-SIT\\\footnotesize ($\downarrow$ better)}} & \multicolumn{1}{c|}{\shortstack{H-SIT\\\footnotesize ($\uparrow$ better)}} &
        \multicolumn{1}{|c}{\shortstack{L-SIT\\\footnotesize ($\downarrow$ better)}} & \multicolumn{1}{c|}{\shortstack{H-SIT\\\footnotesize ($\uparrow$ better)}} &\multicolumn{1}{|c}{\shortstack{L-SIT\\\footnotesize ($\downarrow$ better)}} & \multicolumn{1}{c|}{\shortstack{H-SIT\\\footnotesize ($\uparrow$ better)}}  \\
        \cmidrule(r){1-7} 
        Seg-Sobol       &12.2\%    & 25.9\%    & 12\%    & 29.1\%    &18\%    & 34.1\%   \\
        
        Entropy-Centric & 2\% & 37.4\%      & 0.7\%  & 44\%  &3.4\%         & 48.5\% \\

        Grad-CAM & 27.3\% & 2.2\%      & 29.7\% & -0.1\%  &38.8\%     & 5\%  \\

        Score-CAM & 0.6\% & 32.3\%      & 4.1\% & 36.2\%  &8.9\%        & 45.3\%    \\

        \cmidrule(r){1-7}
    
    \end{tabular}

    \label{tab:Quantitative_Results}
\end{table*}

\begin{table*}[b]
    \caption{Comparative Analysis: Score-CAM vs. Entropy-Centric}
    \centering
    \begin{tabular}{|c|c|c|c|c|c|c|c|c|}
    \hline
    XAI Schemes     & Performance & Stability & \begin{tabular}[c]{@{}c@{}}Background \\ Reliance\end{tabular} & \begin{tabular}[c]{@{}c@{}}Interaction \\ Awareness\end{tabular} & Methodology & Granularity &  \begin{tabular}[c]{@{}c@{}}Primary \\ Objective\end{tabular} \\ \hline
    Score-CAM       &
    +++        &
    Low         &                     Weak          &                   Limited            &         Model-Specific       &
    Target-based            &         
    Fast Visualization
    \\ \hline
    Entropy-Centric &
    ++++ &
    High          &                                                    
    Strong      &                     High       &
    Model-Agnostic   &
    Patch-based            &          
    Model Auditing
    \\ \hline
\end{tabular}
\vspace{0.5em}
\label{tab:horizontal_xai_comparison} 

\end{table*}

\section{Results}
%\begin{highlightparagraph}

This section presents the performance evaluation of the proposed Entropy-Centric XAI method, Grad-CAM, Score-CAM~\cite{gizzini2023extending}, and Seg-Sobol~\cite{fel2021sobol} methods. The XAI evaluation is performed using the L-SIT and the proposed H-SIT XAI evaluation methodology. Moreover, the rooftop U-Net architecture segmentation model \cite{model} trained using the WHU dataset is used as a case study. 

Figure~\ref{fig:heatmaps-results} shows a sample of XAI heatmaps generated by the three methods, Grad-CAM, Score-CAM, and the proposed Entropy-Centric method, from a total of 142 images. This set of images was chosen to reflect the different cases observed in the heatmap results.

Starting with the first image, which represents the majority of the results, Score-CAM and Entropy-Centric produce good explanation heatmaps by highlighting building areas and their surroundings. In contrast, Grad-CAM considers only small portions of the building pixels as important, failing to highlight the complete buildings, not to mention the interaction between the buildings and their surroundings. This behavior generally reflects what Grad-CAM produced across the rest of the test dataset, as can also be observed in the five presented samples.

The second, third, fourth, and fifth images reflect special cases observed with Score-CAM. In some cases, it highlights wide regions far from the buildings, as shown in Images 3 and 4. It may also skip buildings and highlight different objects, such as the road in Image 4, or fail to highlight any significant region, as shown in Image 5. Although these cases appeared only a limited number of times in the test explanations, they represent particular failure cases that are worth mentioning. On the other hand, Entropy-Centric did not exhibit similar failure cases in the examined results. Its least satisfactory cases mainly involved skipping some parts of the buildings without highlighting them.

To summarize the qualitative analysis, unlike Grad-CAM, which highlights only some pixels belonging to the target class, Entropy-Centric and Score-CAM generally provide better heatmaps. The majority of their highly highlighted regions correspond to buildings and their surroundings. These results suggest that building pixels, together with some contextual information from their surroundings, are among the regions that are most relevant to the model’s segmentation predictions.

To further quantitatively evaluate the reliability of the benchmarked XAI schemes, L-SIT and the proposed H-SIT are used in terms of the following metrics: 

\begin{enumerate}[
    label=\textit{(\roman*)},
    leftmargin=*,
    itemsep=0.5em
]

    \item The drop in the model’s prediction confidence inside the target object area compared to the original one when $I$ is passed to the model.

    \item The drop in the Intersection over Union (IoU) score between the perturbed image prediction and the ground truth compared to the IoU score of the original prediction with the ground truth. IoU tells us how much the predicted object overlaps with the correct object.

    \item The increase in the entropy of the target object between the masked-image prediction, using L-SIT and H-SIT, and the original prediction.

\end{enumerate}

Table \ref{tab:Quantitative_Results} shows the quantitative results for the benchmarked XAI methods where the threshold is $0.1$. This threshold allows better evaluation since the majority of the highlighted pixels are included within the heatmap.

For the L-SIT methodology, the low-importance regions defined by the heatmaps below threshold $0.1$ are masked out. As expected, this results in a reduction in prediction confidence and IoU scores, and a rise in entropy, regardless of the method used. Notably, the proposed Entropy-Centric outperforms the Seg-Sobol, Grad-CAM and the Score-CAM methods. This indicates that the Entropy-Centric heatmap more precisely identifies irrelevant areas, as their removal barely impacts model predictions, segmentation accuracy, or confidence. In contrast, Grad-CAM suffers from a considerable performance degradation when low-quality regions are masked, likely due to less discriminative attribution scores that misclassify important pixels as low-impact. 

Concerning the proposed H-SIT XAI evaluation methodology, which focuses on masking high-importance regions as determined by each method, reveals complementary insights. Here, the metrics result in bad performance, reflecting the fact of masking the high-importance regions. In this context, both the proposed Entropy-Centric and the Score-CAM methods consistently yield the largest drops in prediction confidence and IoU, and the greatest increases in entropy, with the superiority of the proposed Entropy-Centric method. This confirms the efficient ability of the proposed Entropy-Centric method to accurately isolate decision-critical regions, making the model highly sensitive to their removal. Seg-Sobol performed moderately with notebly lower performance than Entropy-Centric. We note that the Grad-Cam method barely has any drop in its metrics, which demonstrates its heatmap's deficiency in giving any importance to the regions outside the building areas. This is not always true as segmentation highly relies on the interaction between image regions.

% Its worth to mention that our proposed attribution approach that is based on uncertanity rather than the  

Besides the quantitative performance superiority of the proposed Entropy-Centric over the Score-CAM method, Table \ref{tab:horizontal_xai_comparison} summarizes how both methods differ fundamentally in their computational demands, implementation requirements, and the nature of the explanations they provide for segmentation models. Score-CAM requires a small multiple of a standard forward pass, and produces fine-grained attribution maps whose spatial resolution is tied to the underlying convolutional feature maps. However, this method requires access to internal activation maps of the trained model, making it sensitive to architectural details. In addition, its implementation often relies on model-specific hooks and careful layer selection, which limits portability across architectures. In contrast, the proposed Entropy-Centric operates in a fully black-box manner and relies on systematic input perturbations, resulting in substantially improved stability. Because it does not depend on gradients or internal representations, its explanations remain consistent across architectures and naturally extend to pipelines containing non-differentiable components. Although its explanations are patch-based and therefore limited by the chosen sampling grid, this coarser granularity enables the explicit analysis of contextual dependence and feature interactions by measuring confidence or entropy changes when regions are masked. Consequently, Score-CAM is well-suited for rapid, fine-resolution visualization, whereas the proposed Entropy-Centric method provides more robust and implementation-agnostic explanations that are better aligned with thorough model analysis and verification.

\section{Conclusion}

This work addresses the critical gap in explainable AI methods for image segmentation by proposing an Entropy-Centric XAI method. The proposed method calculates the final relevance scores based on the entropy uncertainty concept. In addition, we propose a robust XAI evaluation methodology denoted as H-SIT. Performance evaluation demonstrates the superior fidelity of the proposed Entropy-Centric method over the benchmarked methods. The proposed Entropy-Centric method is able to effectively isolate irrelevant regions as well as accurately identify decision-critical areas. Therefore, the proposed XAI scheme establishes a robust foundation for XAI methods for semantic segmentation. Future directions include multi-dataset validation, hybrid XAI methods, and diverse architectures support. 

% \newpage
\bibliographystyle{IEEEtranN}
\bibliography{references}

\end{document}